# Phase transition in SONFIS&SORST

Hamed Owladeghaffari

Department of Mining&Metallurgical Engineering
Amirkabir University of Technology
Tehran,Iran

h.o.ghaffari@gmail.com

**Abstract.** In this study, we introduce general frame of MAny Connected Intelligent Particles Systems (MACIPS). Connections and interconnections between particles get a complex behavior of such merely simple system (system in system).Contribution of natural computing, under information granulation theory, are the main topics of this spacious skeleton. Upon this clue, we organize two algorithms involved a few prominent intelligent computing and approximate reasoning methods: self organizing feature map (SOM), Neuro-Fuzzy Inference System and Rough Set Theory (RST). Over this, we show how our algorithms can be taken as a linkage of government-society interaction, where government catches various fashions of behavior: "solid (absolute) or flexible". So, transition of such society, by changing of connectivity parameters (noise) from order to disorder is inferred. Add to this, one may find an indirect mapping among finical systems and eventual market fluctuations with MACIPS.

## 1 Introduction

Complex systems are often coincided with uncertainty and order-disorder transitions. Apart of uncertainty, fluctuations forces due to competition of between constructive particles of system drive the system towards order and disorder. There are numerous examples which their behaviors show such anomalies in their evolution, i.e., physical systems, biological, financial systems [1]. In other view, in monitoring of most complex systems, there are some generic challenges for example sparse essence, conflicts in different levels, inaccuracy and limitation of measurements ,which in beyond of inherent feature of such interacted systems are real obstacle in their analysis and predicating of behaviors. There are many methods to analyzing of systems include many particles that are acting on each other, for example statistical methods [2], Vicsek model [3]. Other solution is finding out of "main nominations of each distinct behavior which may has overlapping, in part, to others". This advance is to bate of some mentioned difficulties that can be concluded in the *"information granules"* proposed by Zadeh [4]. In fact, more complex systems in their natural shape can be described in the sense of networks, which are made of connections among the units. These units are several facets of information granules as well as clusters, groups, communities, modules [5]. Let us consider a more real feature: dynamic natural particles in their inherent properties have (had have-will have) several appearances of "natural" attributes as in individually or in group forms. On the other hand, in society, interacting of main such characteristics (or may extra-natural forces: metaphysic) in facing of predictable or unpredictable events, determines destination of the supposed society.

Based upon the above, hierarchical nature of complex systems [6], developed (developing) several branches of natural computing (and related limbs) [7], collaborations, conflicts, emotions and other features of real complex systems, we propose a general framework of the known computing methods in the connected (or complex hybrid) shape, so that the aim is to inferring of the substantial behaviors of intricate and entangled large societies. Obviously, connections between units of computing cores (intelligent *particles*) can introduce part (or may full) of the comportments (demeanors-deportments...). Complexity of this system, called MAny Connected Intelligent Particles Systems (MACIPS), add to reactions of particles against information flow, and can open new horizons in studying of this big query: *is there a unified theory for the ways in which elements of a system(or aggregation of systems) organize themselves to produce a behavior?*[8].

In this study, we select a very little limited part of MACIPS (fig.1.), as well as hybrid intelligent systems, and investigate several levels of responses in facing of real information. We show how relatively such our simple methods that can produce (mimic) complicated behavior such government-nation interactions. Mutual relations between algorithms layers identify order-disorder transferring of such systems. So, we found our proposed methods have good ability in mimicking of government-nation interactions while government and society can take the different states of responses. Developing of such intelligent hierarchical networks, investigations of their performances on the noisy information and exploration of possible relate between phase transition steps of the MACIPS and flow of information in to such systems are new interesting fields, as well in various fields of science and economy.

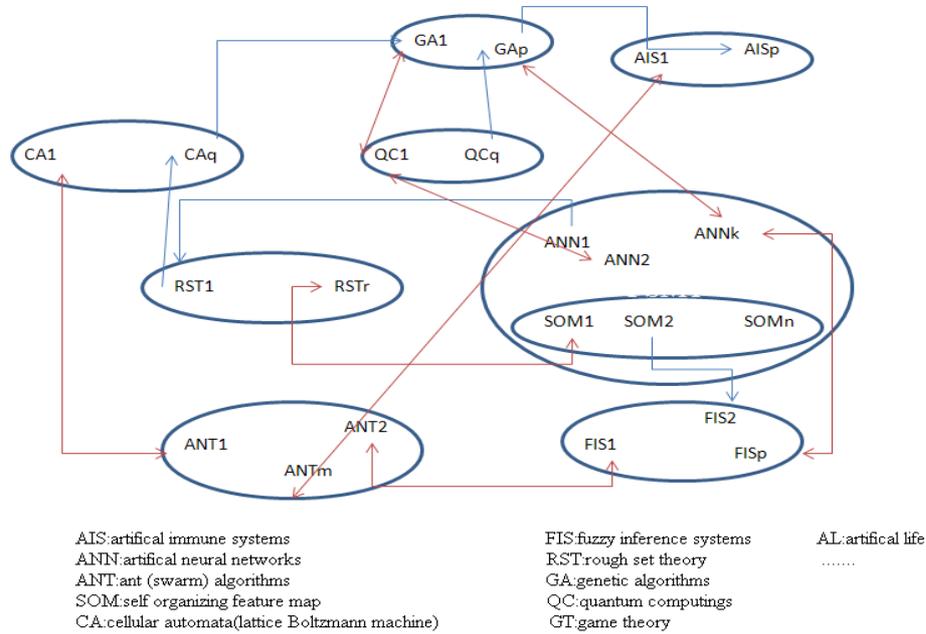

**Fig. 1.** A schematic view of MAny Connected Intelligent Particles Systems (MACIPS)

## 2 instruments

### 2.1 Self Organizing feature Map (SOM)

Kohonen self-organizing networks are competition-based network paradigm for data clustering. The learning procedure of Kohonen feature maps is similar to the competitive learning networks. The main idea behind competitive learning is simple; the winner takes all. The competitive transfer function returns neural outputs of 0 for all neurons except for the winner which receives the highest net input with output 1.
SOM changes all weight vectors of neurons in the near vicinity of the winner neuron towards the input vector. Due to this property SOM, are used to reduce the dimensionality of complex data (data clustering). Competitive layers will automatically learn to classify input vectors, the classes that the competitive layer finds are depend only on the distances between input vectors [9].

### 2.2 Neuro-fuzzy inference system (NFIS)

There are different solutions of fuzzy inference systems. Two well-known fuzzy modeling methods are the Tsukamoto fuzzy model and Takagi– Sugeno–Kang (TSK) model. In the present work, only the TSK model has been considered. The TSK fuzzy inference systems can be easily implanted in the form of a so called Neuro-fuzzy network structure. In this study, we have employed an adaptive neuro-fuzzy inference system [10].

### 2.3 Rough Set Theory (RST)

The rough set theory introduced by Pawlak [11], [12] has often proved to be an excellent mathematical tool for the analysis of a vague description of object. The adjective vague referring to the quality of information means inconsistency, or ambiguity which follows from information granulation.

An information system is a pair $S=< U, A >$, where $U$ is a nonempty finite set called the universe and $A$ is a nonempty finite set of attributes. An attribute $a$ can be regarded as a function from the domain $U$ to some value set $V_a$. An information system can be represented as an attribute-value table, in which rows are labeled by objects of the universe and columns by attributes. With every subset of attributes $B \subseteq A$, one can easily associate an equivalence relation $I_B$ on $U$:

$$I_B = \{(x,y) \in U : for\ every\ a \in B, a(x) = a(y)\} \quad (1)$$

Then, $I_B = \bigcap_{a \in B} I_a$.

If $X \subseteq U$, the sets $\{x \in U : [x]_B \subseteq X\}$ and $\{x \in U : [x]_B \cap X \neq \varphi\}$, where $[x]_B$ denotes the equivalence class of the object $x \in U$ relative to $I_B$, are called the B-lower and the B-upper approximation of X in S and denoted by $\underline{BX}$ and $\overline{BX}$, respectively. Consider $U = \{x_1, x_2, ..., x_n\}$ and $A = \{a_1, a_2, ..., a_n\}$ in the information system $S = \prec U, A \succ$.

By the discernibility matrix $M(S)$ of $S$ is meant an $n*n$ matrix such that

$$c_{ij} = \{a \in A : a(x_i) \neq a(x_j)\} \quad (2)$$

A discernibilty function $f_s$ is a function of m Boolean variables $a_1...a_m$ corresponding to attribute $a_1...a_m$, respectively, and defined as follows:

$$f_s(a_1,...,a_m) = \wedge\{\vee(c_{ij}) : i,j \leq n, j \prec i, c_{ij} \neq \varphi\} \quad (3)$$

Where $\vee(c_{ij})$ is the disjunction of all variables with $a \in c_{ij}$ [13]. Using such discriminant matrix the appropriate rules are elicited. In this study we have developed dependency rule generation –RST- in MatLab7, and on this added toolbox other appropriate algorithms have been prepared.

## 2.4 The proposed procedure

Developed algorithms use four basic axioms upon the balancing of the successive granules assumption:

- Step (1): dividing the monitored data into groups of training and testing data
- Step (2): first granulation (crisp) by SOM or other crisp granulation methods
  Step (2-1): selecting the level of granularity randomly or depend on the obtained error from the NFIS or RST (regular neuron growth)
  Step (2-2): construction of the granules (crisp).
- Step (3): second granulation (fuzzy or rough granules) by NFIS or RST
  Step (3-1): crisp granules as a new data.
  Step (3-2): selecting the level of granularity; (Error level, number of rules, strength threshold...)
  Step (3-3): checking the suitability. (Close-open iteration: referring to the real data and reinspect closed world)
  Step (3-4): construction of fuzzy/rough granules.
- Step (4): extraction of knowledge rules

Selection of initial crisp granules can be supposed as "Close World Assumption (CWA)". But in many applications, the assumption of complete information is not feasible, and only cannot be used. In such cases, an "Open World Assumption (OWA)', where information not known by an agent is assumed to be unknown, is often accepted [14].

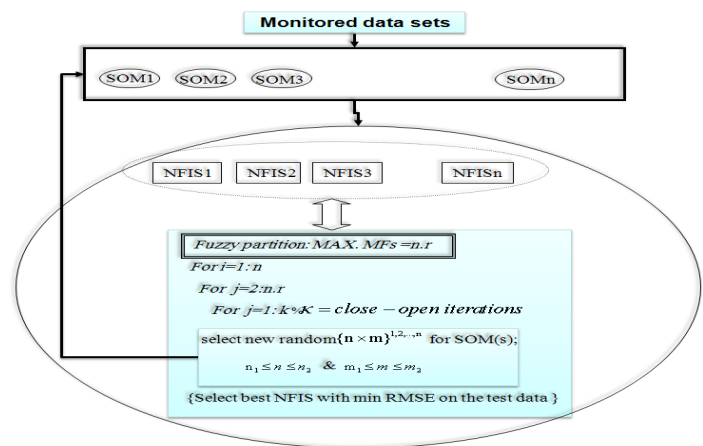

**Fig. 2.** Self Organizing Neuro-Fuzzy Inference System (SONFIS)

Balancing assumption is satisfied by the close-open iterations: this process is a guideline to balancing of crisp and sub fuzzy/rough granules by some random/regular selection of initial granules or other optimal structures and increment of supporting rules (fuzzy partitions or increasing of lower /upper approximations ), gradually. The overall schematic of Self Organizing Neuro-Fuzzy Inference System -Random: SONFIS-R has been shown in Fig.2. In first granulation step, we use a linear relation is given by:

$$N_{t+1} = \alpha N_t + \Delta_t; \Delta_t = \beta E_t + \gamma \quad (4)$$

Where $N_t = n_1 \times n_2; |n_1 - n_2| = Min.$ is number of neurons in SOM or Neuron Growth (NG); $E_t$ is the obtained error (measured error) from second granulation on the test data and coefficients must be determined, depend on the used data set. Obviously, one can employ like manipulation in the rule (second granulation) generation part, i.e., number of rules (as a pliable regulator).

Determination of granulation level is controlled with three main parameters: range of neuron growth, number of rules and error level. The main benefit of this algorithm is to looking for best structure and rules for two known intelligent system, while in independent situations each of them has some appropriate problems such: finding of spurious patterns for the large data sets, extra-time training of NFIS or SOM.

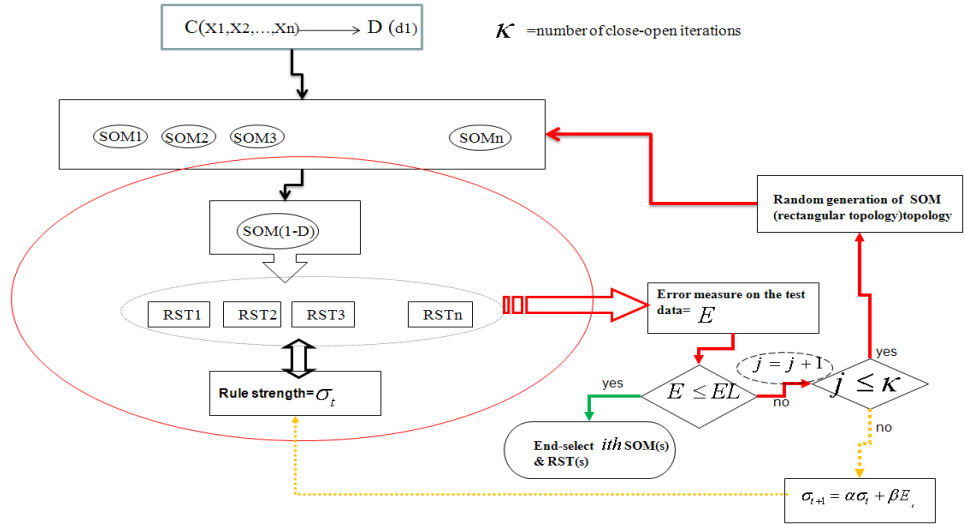

**Fig. 3.** Self Organizing Rough Set Theory-Random neuron growth & adaptive strength factor (SORST-R)

In second algorithm RST instead of NFIS has been proposed (fig. 3). Applying of SOM as a preprocessing step and discretization tool is second process. Categorization of attributes (inputs/outputs) is transferring of the attribute space to the symbolic appropriate attributes. In fact for continuous valued attributes, the feature space needs to be discretized for defining indiscernibilty relations and equivalence classes.

Because of the generated rules by a rough set are coarse and therefore need to be fine-tuned, here, we have used the preprocessing step on data set to crisp granulation by SOM (close world assumption).

In fact, with referring to the instinct of the human, we understand that human being want to states the events in the best simple words, sentences, rules, functions and so forth. Undoubtedly, such granules while satisfies the mentioned axiom that describe the distinguished initial structure(s) of events or immature data sets. Second SOM, as well as close world assumption, gets such dominant structures on the real data.

In other word, condensation of real world and concentration on this space is associated with approximate analysis, such rough or fuzzy facets. Determination of granulation level is controlled with three main parameters: range of neuron growth, number of rules and/or error level. The main benefit of this algorithm is to looking for best structure and rules for two known intelligent system, while in independent situations each of them has some appropriate problems such finding of spurious patterns for the large data sets, extra-time training of NFIS for large data set. So, we can use NFIS as an organizing measurement.

Despite of the aforesaid background behind the proposed algorithms, we can assume interactions of the two layer of algorithm as behaviors of complex systems such: society and government, where reactions of a dynamic community to an "absolute (solid) or flexible" government (regulator) is controlled by correlation factors of the two simplified systems. In absolute case, the second layer (government\ regulator) has limited rules with stable learning iteration for all of matters. In first layer, society selects most main structures of the stimulator where these clusters upon the reaction of government and pervious picked out structures will be adjusted. In facing this, flexible regulator has ability of adapting with the evolution of society.

This situation can be covered by two discrete alternatives: evolution of constitutive rules (policies) over time passing or a general approximation of the dominant rules on the emerged attitudes. In latter case the legislators can considers being conflicts of the emerged states. Other mode can be imagined as poor-revealing structures of the society due to poor-learning or relatively high disturbances within inner layers of the community. It must be noticed, we may choose other two general connected networks or other natural inspired systems involve such hierarchical topology for instances: stock market and stock holders, queen and bees, confliction and quarrel between two countries, interaction among nations (so its outcome can be strategy identifying for trade barriers[19]) and so on. With variations of correlation factors of the two sides (or more connected intelligent particles), one can identify point (or interval) of changes behavior of society or overall system and then controlling of society may be satisfied.

Obviously, considering of simple and reasonable phase transition measure in the mentioned systems will be necessary so that regarding of entropy and order parameter are two distinguished criteria [1].

In this study we use crisp granules level to emulation of phase passing. So, we consider both "absolute and flexible" government while in latter case the approximated rules are contemplated and the presumed society will cope to the disturbances. As before mentioned, in this macroscopic sight of the interactions, the reaction of first layer to the current stimulator is upon the batching inserting of information, and so cannot take in to account of "historical memory decadence" of individuals of the community. In other sense all of information gets an equal effect on the society whereas in actual case "time streamlet bears other generations and new worlds, so new politics".

## 3 phase transition on the "lugeon data set"

In this part of paper, we ensue our algorithms on the "lugeon data set" [15]. This study only considers phase transition view of our proposed algorithms and direct applications of the mentioned systems in other data sets can be found in [15], [16].To evaluate the interactions due to the lugeon values we follow two situations where phase transition measure is upon the crisp granules (here NG): 1) second layer gets a few limited rules by using NFIS; 2) second layer gets all of extracted rules by RST and under an approximated progressing.

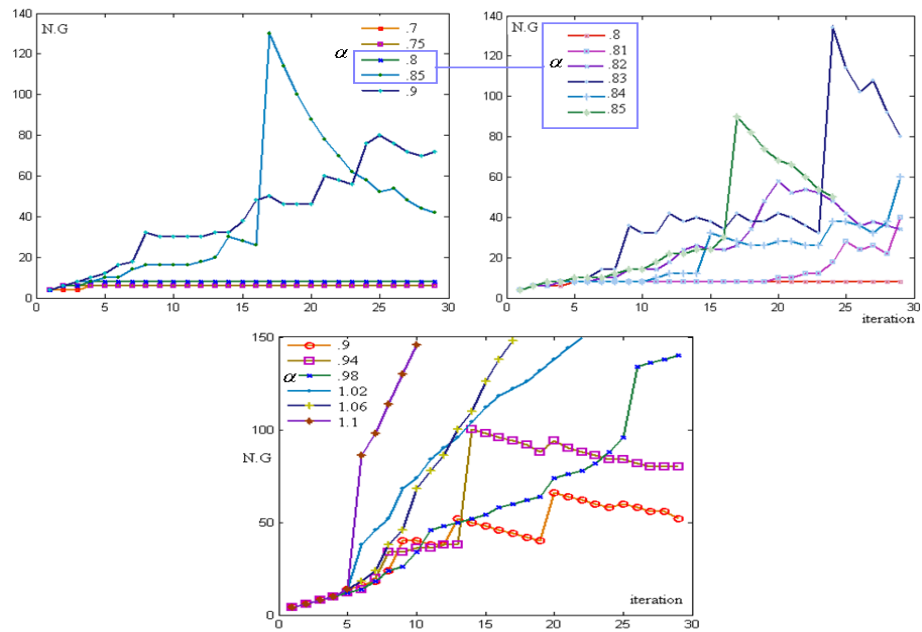

**Fig.4.** Effect of Alpha variations in neuron growth (N.G) of SONFIS-AR with n.r=2 over 30 iterations-; $\beta$ =.001 -a) .7=< Alpha<=.9; b). 8=< Alpha <=.85; c).9=<Alpha<=1.1

Analysis of first situation is started off by setting number of close-open iteration and maximum number of rules equal to 30 and 3 in SONFIS respectively. The error measure criterion in SONFIS is Root Mean Square Error (RMSE), given as below:

$$RMSE = \sqrt{\frac{\sum_{i=1}^{m}(t_i - t_i^*)^2}{m}}$$ ; where $t_i$ is output of SONFIS and $t_i^*$ is real answer; $m$ is

the number of test data(test objects). In the rest of paper, let $m=93$ and number of inserting data set $=600$. By employing of (4) in SONFIS and $\beta$ =.001 and $\gamma$ =.5; the general patterns of NG and RMSE vs. time steps and variations of $\alpha$ can be observed (fig.4). It must be noticed for two like process (i.e., $\alpha$ =.9), we have different situation of neuron growth. The main reason of such behavior is on the regulation of weight neurons in SOM thank to initial random selection and fall in to the "dead neurons state". However, this will be interesting if we see real case, as is appeared in real society, in order to "in an identical cases (but in an unlike iteration) society may shows other behavior, not completely different from other mate".

Fig.4 indicates how the neurons fluctuations with time passing reveal more chaos while the phase transition step can be transpired in $\alpha$ =.8-.85. This evolution of first layer has occurred in a continuous way, as well as other progressing of swarms systems [17]. The comparison between error bar graph and average neuron growth, for this case, exhibits a similar treat on the phase transferring interval (fig.5).

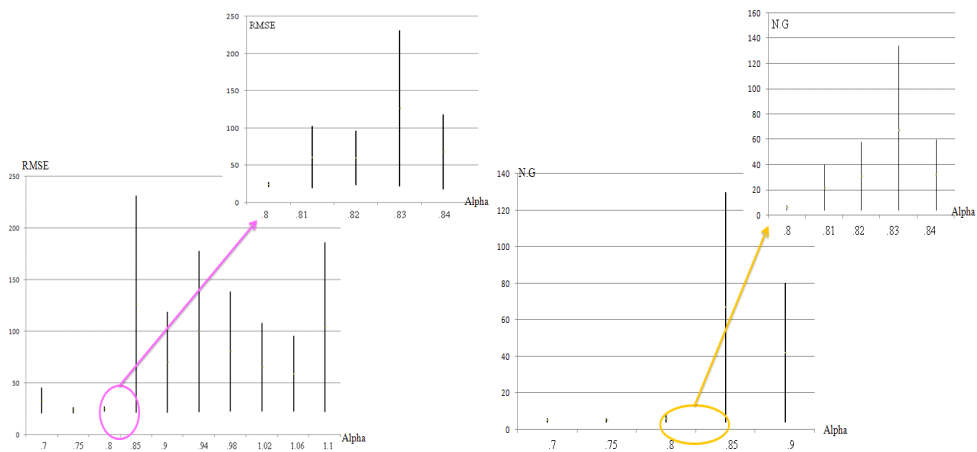

**Fig.5.** Aggregation of Alpha variations-RMSE&N.G in SONFIS with *n.r=2* over 30 iterations (Phase transition step)

With increasing of number of rules (*n.r*) in second layer to 3, our system doesn't disclose a big jump (starting of rush after $\alpha$ =.8) in phase transition step where we preserve previous values for NFIS and an upper limit for N.G, i.e., 150 (fig.6). It must be noticed that the variation of RMSE (or deduction of government from society) is not coincided with N.G whereas with fixing of $\beta$ and changing $\alpha$, the assumed government considers a constant role for him in repression of dynamic society. Let us consider a reverse occasion: $\alpha$ is constant (=.9) and $\beta$ takes different values (fig.7). Such consideration, apart of distinction of the possible phase altering step (after $\beta$ $=4\times10^{-4}$), may display another feature of society alteration: the proper chaos related to the later fashion has larger values so that is not relatively agreed with N.G. In fact, our government loses pervious relative order. In both two former and latter options, the phase transition has been occurred gradationally likewise one can consider three discrete steps to these conversions: society with "silent dead (laminar)", in transition and in triggering of revolutionary community.

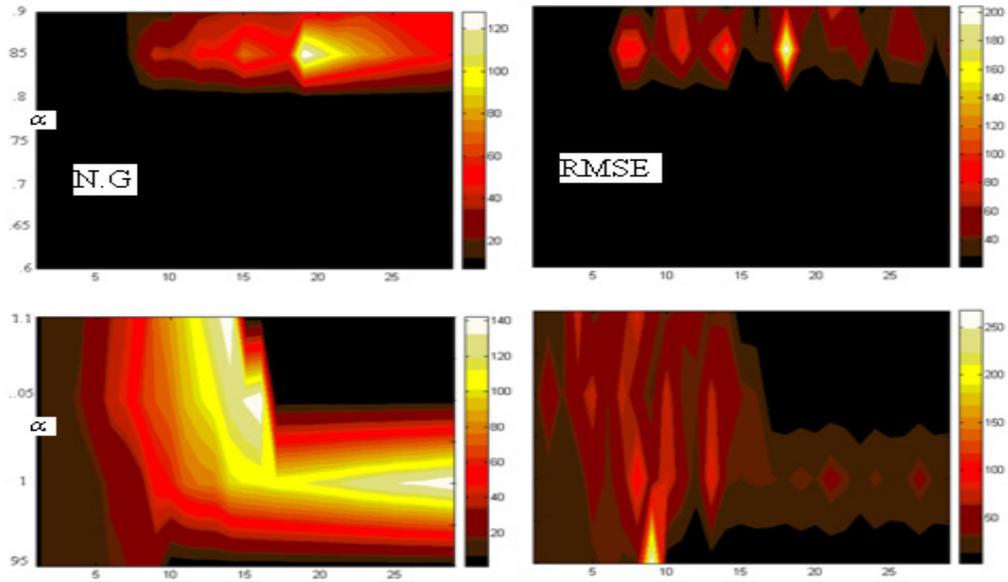

**Fig.6.** Effect of Alpha variations in neuron growth –Beta=.001(N.G)-left- &RMSE-right- of SONFIS with n.r=3 over 30 iterations; a) .6=< Alpha<=.9; b) .95=< Alpha <=1.1

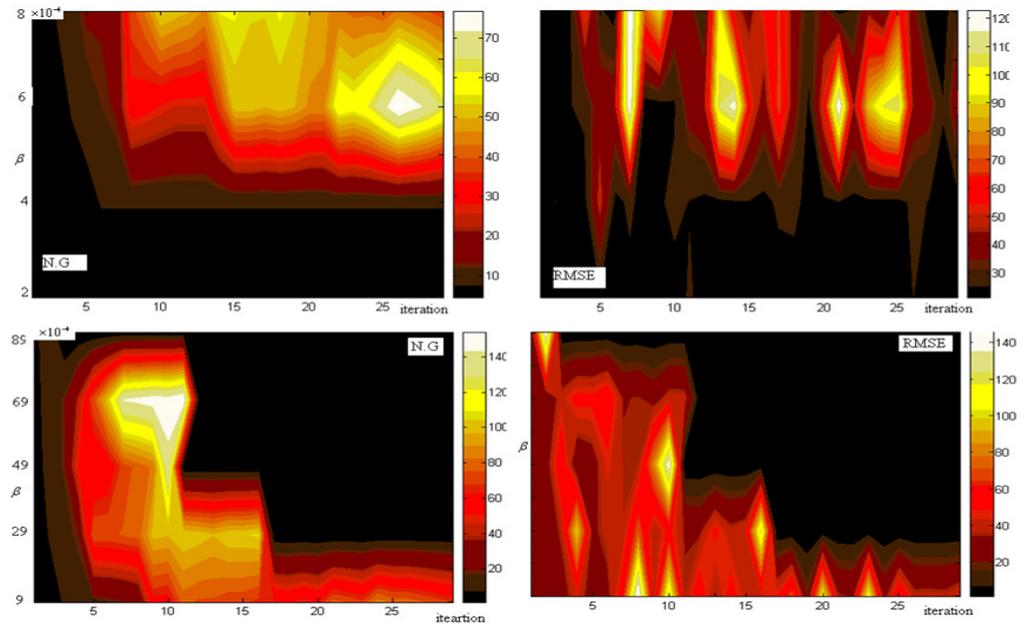

**Fig.7.** Effect of Beta variations-Alpha =.9 - in neuron growth (N.G)-left- &RMSE-right- of SONFIS-AR with n.r=2 over 30 iterations; a) $2*10^{-4}$=< Beta<=$8*10^{-4}$; b) $9*10^{-4}$=< Beta<=$85*10^{-4}$

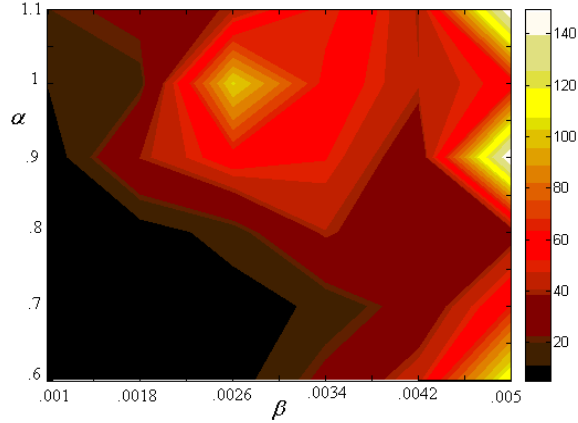

**Fig.8.** Effect of Alpha & Beta variations over 10 iterations on the average NG-*n.r=2*

Now, the possible question may come in to the mind how the sensitivity of first layer to those parameters is. To answer this query, in other process, simultaneously $\alpha$ and $\beta$ are changed over 10 iterations (fig.8.). In a scaled coordination, it can be probed that the rate of $\alpha/\beta$ for the lugeon data set and for laminar state is near to 1, even if $\alpha$ takes a little bigger value. This feature expresses response of overall system to the current wave of data, is more depends on the pervious state of society than to the government reaction. Another interesting result of such accomplishing is that the behavior of the system in large values of $\alpha$ and $\beta$: in despite of some anomalies, for large digits system doesn't consider other disorder parameter and fall in to the disordering way.

In second situation instead of NFIS, we employ RST upon this assumption that the government based on history, experience and other like fashions in the world, can has ability to elicitation of relatively approximated rules of the observed and distinguished behaviors (by transferring of attributes to three scaled classes using 1-D SOM, as well as low, middle and high). The applied error measure for measure of performance of RST is given by: $MSE = \dfrac{\sum_{i=1}^{m}(d_i^{real} - d_i^{classified})^2}{m}$ ;

In deducing of decision for approximated rules (not unique decision part), we select highest value (largest ambiguities) for such decisions. By repeating of steps as well pervious situation, we obtain other behavior of SORST where we employ $\gamma = 1$, in equation 4(fig.9&fig.10). Fig.9 shows how with keeping of $\beta$ as a fixed value, N.G after $\alpha = .8$ gets in to the steeper rate while it has endured relatively high transition time. In fact government without changing of his affecting power ($\beta$) preserve, in a long time society between "tranquil and rushing". Other guessed difference with first option, is on the fast change of the society over passing time and for high $\alpha$ values as though MSE exhibits a low range of variations.

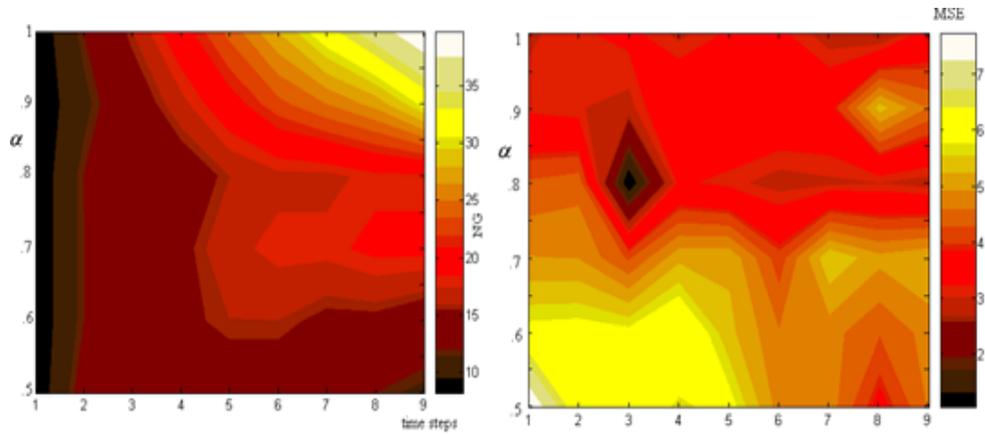

**Fig.9.** Color code of alpha variations in SORST on the N.G and MSE over 9 time steps- $\beta$ =1.01

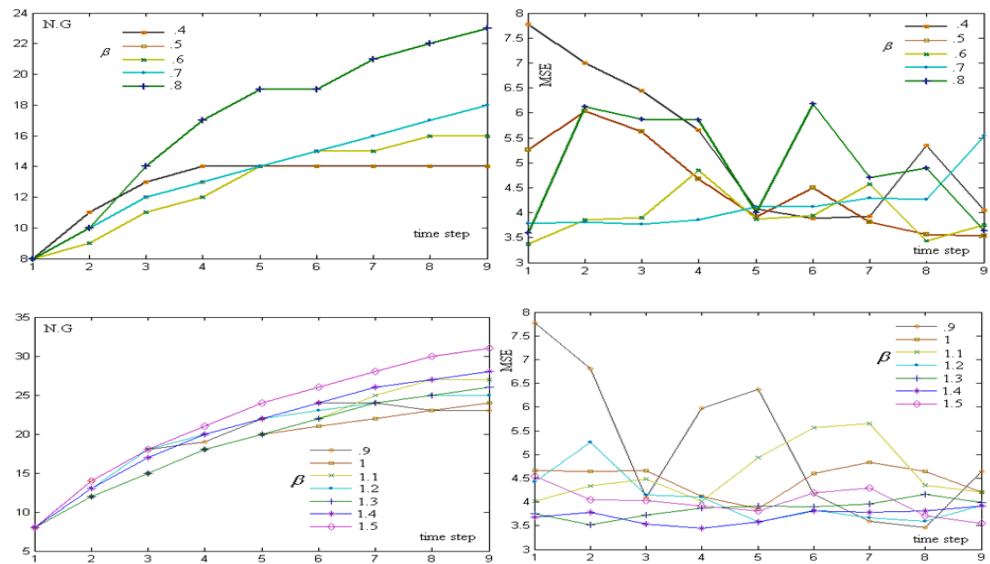

**Fig.10.** Effect of $\beta$ variations in neuron growth (N.G) &MSE of SORST over 9 iterations-; $\alpha$ =.8

We can investigate other more complex cases, for example in [18]; we have explained how the rate of scaling of attributes displays other phase transition state. One may investigate the being of elate among entropy of actual data set and transitions points. Answering to similar questions can instill crucial ideas in complex systems, i.e., how real disturb wave can be controlled by government (or other regulator), is will be necessary to modifying of disorders factors, what will be the extra-forces ($\gamma$) influences.

## 4  conclusions

In this study we proposed two new algorithms in which SOM, NFIS and RST, based on general frame of MAny Connected Intelligent Particles Systems (MACIPS), make SONFIS and SORST. Main idea behind our algorithms is to finding out of best reduced objects, are in balance with second granulation level. Mutual relations between algorithms layers identify order-disorder transferring of such systems. So, we found our proposed methods have good ability in mimicking of government-nation interactions while government and society can take the different states of responses. Developing of such intelligent hierarchical networks, investigations of their performances on the noisy information and exploration of possible relate between phase transition steps of the MACIPS and flow of information in to such systems are new interesting fields, as well in various fields of science and economy.